\documentclass{llncs}
\usepackage[german,english]{babel}
\usepackage{times,theorem}
\usepackage{latexsym,color,graphics}
\usepackage{diagrams}
\diagramstyle[tight,centredisplay%
,dpi=600
,UglyObsolete
,heads=LaTeX%
]

\usepackage{epsfig}
\usepackage{amssymb}
\usepackage{amsmath}
\usepackage{amsfonts}
\usepackage{xspace}
\usepackage{graphicx}
\begingroup
\catcode`\~=11
\gdef\urltilde{\lower 0.6ex\hbox{~}}
\endgroup

\newcommand{\A}{\mathcal{A}} 
 \newcommand{\D}{\mathcal{D}}
\newcommand{\E}{\mathcal{E}}

\newcommand{\K}{\mathcal{K}} \renewcommand{\L}{\mathcal{L}}
 \newcommand{\N}{\mathcal{N}}
 \renewcommand{\P}{\mathcal{P}}
 
 \newcommand{\T}{\mathcal{T}}
 \newcommand{\V}{\mathcal{V}}
\newcommand{\W}{\mathcal{W}}

\title{Strong-AI Autoepistemic Robots Build on Intensional First Order Logic}
\author{Zoran Majki\'c}
\authorrunning{Zoran Majki\'c}
\institute{ISRST, Tallahassee, FL, USA\\
\email{majk.1234@yahoo.com}\\ http://zoranmajkic.webs.com/}

\begin{document}

\maketitle

\begin{abstract} Neuro-symbolic AI attempts to integrate neural and symbolic architectures in a manner that addresses strengths and weaknesses of each, in a complementary fashion, in order to support robust strong AI capable of reasoning, learning, and cognitive modeling. In this paper we consider the
 intensional First Order Logic (IFOL) \cite{Majk22} as a symbolic architecture of modern robots, able to use natural languages to communicate with humans and to reason about their own knowledge with self-reference and abstraction language property.\\
  We intend to obtain the grounding of robot's language by experience of how it uses its neuronal architectures and hence by associating this experience with the mining (sense) of non-defined language concepts (particulars/individuals and universals) in PRP (Properties/Relations/Propositions) theory of IFOL.\\ We consider the  robot's  four-levels knowledge structure: The syntax level of particular natural language (Italian, French, etc..), two universal language levels: its semantic logic structure (based on virtual predicates of FOL and logic connectives), and its corresponding conceptual PRP structure level which universally represents the composite mining of FOL formulae grounded on the last robot's neuro-system level.\\
  Finally, we provide the general method how to implement in IFOL (by using the abstracted terms) different kinds of modal logic operators and their deductive axioms: we present a particular example of robots autoepistemic deduction capabilities by introduction of the special temporal $Konow$ predicate and deductive axioms for it: reflexive, positive introspection and distributive axiom.
\end{abstract}

\section{Introduction}

 The central hypothesis of cognitive science is that thinking can best be understood in terms of representational structures in the mind and computational procedures that operate on those structures. Most work in cognitive science assumes that the mind has mental representations analogous to computer data structures, and computational procedures similar to computational algorithms.

 Main stream machine learning research on deep artificial neural networks  may even be characterized as being behavioristic. In contrast, various sources of evidence from cognitive science suggest that human brains engage in the active development of compositional generative predictive models  from their self-generated sensorimotor experiences. Guided by evolutionarily-shaped inductive learning and information processing biases, they exhibit the tendency to organize the gathered experiences into event-predictive encodings. Meanwhile, they infer and optimize behavior and attention by means of both epistemic- and homeostasis-oriented drives.

  \emph{Knowledge representation}, strongly connected to the problem if knowledge processing, reasoning and "drawing inferences", is one of the main topics in AI. By reviewing the knowledge representation techniques that have been used by humans we will be aware of the \emph{importance of language}. The predominant part of IT industry and user's applications is based on some sublanguage of the standard (extensional) FOL (First Order Logic) with Tarski's semantics based (only) on the truth; my effort is to pass to a more \emph{powerful evolution of the FOL} able to support the meaning of knowledge as well,  by replacing the standard FOL and its  DB theory and practice in IT business.
 All this work is summarized and extended also to AI applications of many-valued logics in my recent book \cite{Majk22}.

 Last 15 years of my work in AI was mainly dedicated to development of a new intensional FOL, by integrating  Montague's and algebraic Bealer's \cite{Beal82} approaches, with a conservative Tarski's semantics of the standard FOL. Basic result was the publication of the conservative extension of Tarski's semantics to intensional FOL \cite{Majk12a}, and two-step intensional semantics \cite{Majk11TS}, which  guaranteed a conservative extension of current RDB, but more than 50-years old technology, toward new IRDB (Intensional RDB). Indeed, in my next Manifesto of IRDB \cite{Majk14R}, I hoped also to find interested research groups and funds to begin the realization of IRDB as a new platform (compatible with all previously developed RDB application), able also to support NewSQL for Big Data, and ready for other AI improvements.

 This paper is an extension (by Section 4) of the paper \cite{Majk23r}.
 It  is dedicated to show how this defined IFOL in \cite{Majk22} can be used for a new generation of intelligent robots, able to communicate with humans with this intensional FOL supporting the meaning of the words and their language compositions. As in \cite{Jack83} we can consider three natural language levels: The \emph{syntax} of a particular natural language (French, English, etc..) its \emph{semantic logic structure} (transformation of parts of the language sentences into the logic predicates and definition of corresponding  FOL formulae) and its corresponding \emph{conceptual structure}, which differently from the semantic layer that represents only the logic's semantics, represents the composed meaning of FOL formulae based on the grounding of intensional PRP concepts.

 Thus, intensional mapping from the free FOL syntax algebra into the algebra of intensional PRP concepts, $$I:\A_{FOL}\rightarrow \A_{int}$$ provided by IFOL theory, is a part of the semantics-conceptual mapping of natural languages. Note that differently from the particularity of any given natural language of humans, the underlying logical semantics and conceptual levels have universal human knowledge structure, provided by innate human brain structure able to rapidly acquire the ability to use any natural language.

 Parsing, tokenizing, spelling correction, part-of-speech tagging, noun and verb phrase chunking are all aspects of natural language processing long handled by symbolic AI, and has to be improved by deep learning approaches. In symbolic AI, discourse representation theory and first-order logic have been used to represent sentence meanings.
 We consider that the natural language (first level) can be parsed into a logical FOL formula with a numbers of virtual predicates and logic connectives of the FOL.
 By such a parsing we obtain the second, semantic logic, structure corresponding to some FOL formula.
 However, natural language is grounded in experience. Humans do not always define all words in terms of other words, humans understand many basic words in terms of associations with sensory-motor experiences for example. People must interact physically with their world to grasp the essence of words like "blue," "could," and "left." Abstract words are acquired only in relation to more concretely grounded terms.

 Theoretical neuroscience is the attempt to develop mathematical and computational theories and models of the structures and processes of the brains of humans and other animals. If progress in theoretical neuroscience continues, it should become possible to tie psychological to neurological explanations by showing how mental representations such as \emph{concepts} are constituted by activities in neural populations, and how computational procedures such as spreading activation among concepts are carried out by neural processes. Concepts, which partly correspond to the words in spoken and written language, are an important kind of mental representation.

 Alan Turing developed the Turing Test in 1950  in his paper, "Computing Machinery and Intelligence". Originally known as the Imitation Game, the test evaluates if a machine’s behavior can be distinguished from a human. In this test, there is a person known as the "interrogator" who seeks to identify a difference between computer-generated output and human-generated ones through a series of questions. If the interrogator cannot reliably discern the machines from human subjects, the machine passes the test. However, if the evaluator can identify the human responses correctly, then this eliminates the machine from being categorized as intelligent.

 Differently from the \emph{simulation} of AI by such Turing tests and the Loebner Prize\footnote{The Loebner Prize was an annual competition in artificial intelligence that awards prizes to the computer programs considered by the judges to be the most human-like. The prize is reported as defunct since 2020.[1] The format of the competition was that of a standard Turing test. In each round, a human judge simultaneously holds textual conversations with a computer program and a human being via computer. Based upon the responses, the judge must decide which is which.} and in accordance with Marvin Minsky\footnote{In the early 1970s, at the MIT Artificial Intelligence Lab, Minsky and Papert started developing what came to be known as the Society of Mind theory. The theory attempts to explain how what we call intelligence could be a product of the interaction of non-intelligent parts. Minsky says that the biggest source of ideas about the theory came from his work in trying to create a machine that uses a robotic arm, a video camera, and a computer to build with children's blocks. In 1986, Minsky published The Society of Mind, a comprehensive book on the theory which, unlike most of his previously published work, was written for the general public.\\
 In November 2006, Minsky published The Emotion Machine, a book that critiques many popular theories of how human minds work and suggests alternative theories, often replacing simple ideas with more complex ones.
}, in this paper I argue that a real AI for robots can be obtained by using formal Intensional FOL (with defined intensional algebra of intensions of language constructions) for the robots as their symbolic AI component, by defining the sense to ground terms (the words) in an analog way, associating to these words the software processes developed for the robots when they recognize by these algorithms (neural architectures) the color "blue" of visual objects, the position "left" etc... In this way we would obtain a \emph{neuro-symbolic AI} which attempts to integrate neural and symbolic architectures in a manner that addresses strengths and weaknesses of each, in a complementary fashion, in order to support robust AI capable of reasoning, learning, and cognitive modeling. To build a robust, knowledge-driven approach to AI we must have the machinery of symbol-manipulation as, in this case, an IFOL. Too much of useful knowledge is abstract to make do without tools that represent and manipulate abstraction, and to date, the only machinery that we know of that can manipulate such abstract knowledge reliably is the apparatus of symbol-manipulation. The IFOL defined in \cite{Majk22} is provided by abstraction operators as well.

Daniel Kahneman \cite{Kahn11} describes human thinking as having two components, System 1 and System 2. System 1 is fast, automatic, intuitive and unconscious. System 2 is slower, step-by-step, and explicit. System 1 is the kind used for pattern recognition while System 2, in uor case based on IFOL, is far better suited for planning, deduction, and deliberative thinking. In this view, deep learning best models the first kind of thinking while symbolic reasoning best models the second kind and both are needed.

So, for the words (ground linguistic terms), which can not be "defined by other words", the robots would have some own internal experience of the concrete sense of them. Thus, by using intensional FOL the robots can formalize also the natural language expressions "I see the blue color" by a predicate "see(I,blue color)" where the sense of the ground term "I" (\emph{Self})\footnote{Self in a sense which implies that all our activities are controlled by powerful creatures inside ourselves, who do our thinking and feeling for us.} for a robot is the name of the  main working coordination program which activate all other algorithms (neuro-symbolic AI subprograms) like visual recognition of color of the object in focus. But also the auto-conscience sentence like "I know that I see the blue color" by using abstracting operators "$\lessdot\_\gtrdot$" of intensional FOL, expressed by the predicate "know(I,$\lessdot$ see(I, blue color)$\gtrdot$)", etc...

Consequently, we argue that by using this intensional FOL, the robots can develop their own knowledge about their experiences and communicate by a natural language with humans. So, we would be able to develop the interactive robots which learn and understand spoken language via multisensory grounding and internal robotic embodiment.

The \emph{grounding} of the intensional concepts i PRP theory of intensional logic was not considered in my recent book \cite{Majk22} from the fact that this book was only restricted on the symbolic AI aspects (IFOL); so by this paper we extend the logic theory developed in \cite{Majk22} with concrete grounding of its intensional concepts in order to obtain a strong AI for robots. So, in next Section we will provide a short introduction to IFOL  and its intensional/extensional semantics \cite{Majk22}.

\section{Algebra for Composition of Meanings in IFOL}
Contemporary use of the term "intension" derives from the
traditional logical doctrine that an idea has both an extension and
an intension. Although there is divergence in formulation, it is
accepted that the extension of an idea  consists of the subjects to
which the idea applies, and the intension consists of the attributes
implied by the idea. In contemporary philosophy, it is linguistic
expressions (here it is a logic formula), rather than concepts, that
are said to have intensions and extensions. The intension is the
concept expressed by an expression of intensional algebra $\A_{int}$, and the extension is the set of items to which the expression applies. This usage resembles use of Frege's use  of "Bedeutung" and "Sinn" \cite{Freg92}.

Intensional entities (or  concepts) are such things as Propositions,
Relations and Properties (PRP). What make them "intensional" is that they
violate the principle of extensionality; the principle that
extensional equivalence implies identity. All (or most) of these
intensional entities have been classified at one time or another as
kinds of Universals \cite{Beal93}.

In a predicate logics, (virtual) predicates  expresses classes
(properties and relations), and sentences express propositions. Note
that classes (intensional entities) are \emph{reified}, i.e., they
belong to the same domain as individual objects (particulars). This
endows the intensional logics with a great deal of uniformity,
making it possible to manipulate classes and individual objects in
the same language. In particular, when viewed as an individual
object, a class can be a member of another class.
\index{virtual predicates}
\begin{definition} \label{def:virt-predicate} \textsc{Virtual predicates:}
\emph{Virtual predicate} obtained from an open formula $\phi \in \L$
is denoted by $\phi(x_1,...,x_m)$ where $(x_1,...,x_m)$ is a particular fixed sequence of the set of all free variables in $\phi$. This definition contains the precise method of establishing the \emph{ordering} of variables in this tuple:
such an method that will be adopted here is the ordering of appearance, from left to right, of free variables in $\phi$. This method of composing the tuple of free variables is  unique and canonical way of definition of the virtual predicate from a given open formula.

The virtual predicates are useful also to replace the general FOL quantifier on variables $(\exists x)$ by specific quantifiers $\exists_i$ of the FOL syntax algebra $\A_{FOL}$, where $i\geq 1$ is the position of variable $x$ inside a virtual predicate. For example, the standard FOL formula $(\exists x_k) \phi(x_i,x_j,x_k,x_l,x_m)$ will be mapped
into intensional concept $\exists_3 \phi(\textbf{x}) \in \A_{FOL}$ where $\textbf{x}$ is the list(tuple) of variables $(x_i,x_j,x_k,x_l,x_m)$.
\end{definition}
Virtual predicates are atoms used to build the \emph{semantic logic structures} of logic-semantics level of any given natural language.

Let us define the FOL syntax algebra $\A_{FOL}$.\\
 For example, the FOL formula
$\phi(x_i,x_j,x_k,x_l,x_m) \wedge \psi (x_l,y_i,x_j,y_j)$ will be
replaced by a specific \emph{virtual predicate} $\phi(x_i,x_j,x_k,x_l,x_m)
\wedge_S \psi~ (x_l,y_i,x_j,y_j)$, with the set of joined variables (their positions in the first and second virtual predicate, respectively) $S = \{(4,1),(2,3)\}$, so that its extension is expressed by an algebraic expression $~R_1 \bowtie_{S}R_2$, where $R_1, R_2$ are the extensions for a given Tarski's interpretation $I_T$ of the virtual predicate $\phi, \psi$ relatively, and the binary operator $~\bowtie_{S}$  is the natural join of these two relations.
  In this example the resulting relation will have the following ordering of attributes: $(x_i,x_j,x_k,x_l,x_m,y_i,y_j)$.
In the case when $S$ is empty (i.e. its cardinality $|S| =0$) then the resulting relation is the Cartesian product of $R_1$ and $R_2$.
For the existential quantification, the FOL formula $(\exists x_k) \phi(x_i,x_j,x_k,x_l,x_m)$ will be replaced  in $\A_{FOL}$ by a specific virtual predicate $(\exists_3) \phi(x_i,x_j,x_k,x_l,x_m)$. For logic negation operator we will use the standard symbol $\neg$.

Based on the new set of logical connectives introduced above, where the  standard FOL operators $\wedge$ and $\exists$ are substituted by a set of specialized operators $\{\wedge_S\}_{S \in \P(\mathbb{N}^2)}$ and $\{\exists n\}_{n \in
\mathbb{N}}$ as explained above, we can define the following free syntax algebra for the FOL:\index{FOL syntax algebra}
 \begin{definition} \label{coro:intensemant} \textsc{FOL sintax algebra:}\\
 Let $\A_{FOL} = (\L, \doteq, \top, \{\wedge_S\}_{S \in
\P(\mathbb{N}^2)}, \neg, \{\exists n\}_{n \in \mathbb{N}})$ be an
\emph{extended}  free syntax algebra for the First-order logic with identity $\doteq$, with the set $\L$ of first-order logic formulae  with the set of variables in $\V$, with $\top$ denoting the tautology formula (the contradiction formula is denoted by $\bot \equiv \neg \top$).
\end{definition}
 We begin with the informal theory that universals (properties (unary relations),
relations, and propositions in PRP theory \cite{Beal79}) \index{PRP theory} are genuine entities that bear fundamental logical relations to one another. To
study properties, relations and propositions, one defines a family
of set-theoretical structures, one define the intensional algebra, a
family of set-theoretical structures most of which are built up from
arbitrary objects and fundamental logical operations (conjunction,
negation, existential generalization,etc..) on them.
\begin{definition} \label{def:Idomain} \textsc{Intensional logic PRP domain $\D$:}\\
In intensionl logic the concepts (properties, relations and
propositions) are denotations for open and closed logic sentences,
thus elements of the structured domain   $~\D = D_{-1} + D_I$, (here
$+$ is a disjoint union) where
\begin{itemize}
  \item A subdomain $D_{-1}$ is made of
 particulars (individuals).
  \item The rest $D_I = D_0 +
 D_1 ...+ D_n ...$ is made of
 universals (\emph{concepts})\footnote{In what follows we will define also a language of concepts with intensional connectives defined as operators of the intensional algebra $\A_{int}$ in Definition \ref{def:intalgebra}, so that $D_I$ is the set of terms of this intensional algebra. }: $D_0$ for  propositions with a distinct concept $Truth \in D_0$, $D_1$ for properties
 (unary concepts)  and  $D_n, n \geq 2,$ for n-ary concept.
\end{itemize}
\end{definition}
  The  concepts in $\D_I$ are denoted by $u,v,...$, while the
 values (individuals) in $D_{-1}$ by $a,b,...$ The empty tuple $<>$ of the nullary relation $r_\emptyset$ (i.e. the unique tuple of 0-ary relation) is an individual in  $D_{-1}$, with $\D^0 =_{def} \{<>\}$. Thus, we have that
  $\{f,t\} = \P(\D^0) \subseteq \P(D_{-1})$, where by $f$ and $t$
 we denote  the empty set $\emptyset$ and set $\{<>\}$ respectively.

The \emph{intensional interpretation} is a mapping between the set $\L$ of formulae of the FOL  and  intensional entities in $\D$, $I:\L \rightarrow \D$, is a kind of  "conceptualization", such that  an open-sentence (virtual predicate)
 $\phi(x_1,...,x_k)$ with a tuple of all free variables $(x_1,...,x_k)$ is mapped into a k-ary \emph{concept}, that is, an intensional entity  $u = I(\phi(x_1,...,x_k)) \in D_k$, and (closed) sentence $\psi$ into a proposition (i.e., \emph{logic} concept) $v =  I(\psi) \in D_0$ with $I(\top) = Truth \in D_0$ for the FOL tautology $\top \in \L$  (the falsity in the FOL is a logic formula $\neg\top \in \L$). A language constant $c$ is mapped into a particular $~a \in D_{-1}$  (intension of $c$) if it is a proper name, otherwise in a correspondent concept $u$ in $D_I$. Thus, in any application of intensional FOL, this intensional interpretation that determines the meaning (sense) of the knowledge expressed by logic formulae is \emph{uniquely determined (prefixed)} (for example, by a grounding on robot's neuro system processes, explained in next section).

 However, the extensions of the concepts (with this prefixed meaning) vary from a context (possible world, expressed by an extensionalizzation function) to  another context in a similar way as for different Tarski's interpretations of the FOL:
\begin{definition} \label{def:extent} \textsc{Extensions and extensionalization functions:}\\
 Let $~\mathfrak{R} = \bigcup_{k \in \mathbb{N}} \P(\D^k) = \sum_{k\in \mathbb{N}}\P(D^k)$ be the set of all k-ary relations, where $k \in \mathbb{N} =
\{0,1,2,...\}$. Notice that $\{f,t\} = \P(\D^0) \subseteq \mathfrak{R}$,
that is, $f,t \in\mathfrak{R}$ and hence the truth values are extensions in $\mathfrak{R}$.

We define the function $f_{<>}:\mathfrak{R}\rightarrow \mathfrak{R}$, such that for any $R \in \mathfrak{R}$,
\begin{equation} \label{eq:true-fal}
f_{<>}(R) =_{def} \{<>\}~ \emph{if} ~R \neq \emptyset;~~ \emptyset~ \emph{otherwise}
\end{equation}
 The extensions of the intensional entities (concepts) are given by the set
 $\E$ of extensionalization functions $h:\D \rightarrow
 D_{-1} +\mathfrak{R}$, such that
 \begin{equation}\label{eq:ExtenFunct}
 h = h_{-1} + h_0 + \sum_{i\geq 1}h_i:\sum_{i
\geq -1}D_i \longrightarrow D_{-1} + \{f,t\} + \sum_{i\geq
1}\P(D^i)
\end{equation}
 where $h_{-1}:D_{-1} \rightarrow D_{-1}$ for the particulars, while
$~h_0:D_0 \rightarrow \{f,t\} = \P(\D^0)$ assigns the truth values
in $ \{f,t\}$ to all propositions with the constant assignment
$h_0(Truth) = t = \{<>\}$, and for each $i\geq 1$, $h_i:D_i \rightarrow \P(D^i)$
assigns a relation to each concept.

Consequently, intensions can be seen as \verb"names" (labels) of atomic or
composite concepts, while the extensions correspond to various rules that
these concepts play in different worlds.
\end{definition}
The intensional entities for the same logic formula, for example $x_2 +3 = x_1^2-4$, which can be denoted by $\phi(x_2,x_1)$ or $\phi(x_1,x_2)$, from above we need to differentiate their concepts by $I(\phi(x_2,x_1)) \neq I(\phi(x_1,x_2))$ because otherwise we would obtain erroneously that $h(I(\phi(x_2,x_1))) = h(I(\phi(x_1,x_2)))$. Thus, in intensional logic the ordering in the tuple of variables $\textbf{x}$ in a given open formula $\phi$ is very important,
and explains why we introduced in FOL the virtual predicates in Definition  \ref{def:virt-predicate}.
\begin{definition} \label{def:relatOperat}
Let us define the extensional relational algebra  for the FOL by, \index{FOL extensional algebra}

$\A_{\mathfrak{R}} = (\mathfrak{R}, R_=, \{<>\}, \{\bowtie_{S}\}_{ S
\in \P(\mathbb{N}^2)}, \sim, \{\pi_{-n}\}_{n \in \mathbb{N}})$,
\\where $ \{<>\} \in \mathfrak{R}$ is the algebraic value
correspondent to the logic truth, $R_=$ is the binary relation
for extensionally equal elements, with the following operators:
\begin{enumerate}
\item Binary operator $~\bowtie_{S}:\mathfrak{R} \times \mathfrak{R} \rightarrow
\mathfrak{R}$,
 such that for any two relations $R_1, R_2 \in
 \mathfrak{R}~$, the
 $~R_1 \bowtie_{S} R_2$ is equal
to the relation obtained by natural join
 of these two relations $~$ \verb"if"
 $S$ is a non empty
set of pairs of joined columns of respective relations (where the
first argument is the column index of the relation $R_1$ while the
second argument is the column index of the joined column of the
relation $R_2$); \verb"otherwise" it is equal to the cartesian
product $R_1\times R_2$.
\item Unary operator $~ \sim:\mathfrak{R} \rightarrow \mathfrak{R}$, such that for any k-ary (with $k \geq 1$) relation $R \in  \P(\D^{k}) \subset \mathfrak{R}$
 we have that $~ \sim(R) = \D^k \backslash R \in \P(\D^{k})$, where '$\backslash$' is the substraction of relations. For $u\in \{f,t\} = \P(\D^0) \subseteq \mathfrak{R}$, $~ \sim(u) = \D^0 \backslash u$.
\item Unary operator $~ \pi_{-n}:\mathfrak{R} \rightarrow \mathfrak{R}$, such that for any k-ary (with $k \geq 1$) relation $R \in \P(\D^{k}) \subset \mathfrak{R}$ we have that $~ \pi_{-n} (R)$ is equal to the relation obtained by
elimination of the n-th column of the relation $R~$ \verb"if" $1\leq
n \leq k$ and $k \geq 2$; equal to, from (\ref{eq:true-fal}), $~f_{<>}(R)~$ \verb"if" $n = k =1$; \verb"otherwise" it is equal to $R$.
\end{enumerate}
We will use the symbol '$=$' for the extensional identity for relations in $\mathfrak{R}$.
\end{definition}
The intensional semantics of the logic language with the set of formulae
$\L$ can be represented by the  mapping
\begin{center}
$~ \L ~\longrightarrow_I~ \D ~\Longrightarrow_{h \in \E}~\mathfrak{R}$,
\end{center}
where $~\longrightarrow_I~$ is a \emph{fixed intensional} interpretation $I:\L \rightarrow \D$ with image $im(I)\subset \D$, and $~\Longrightarrow_{h \in
\E}~$ is \emph{the set} of all extensionalization functions $h:im(I) \rightarrow D_{-1} +\mathfrak{R}$ in $\E$.

 So, we can define only the minimal intensional algebra (with minimal
number of operators) $\A_{int}$ of concepts, able to support the homomorphic extension $$h:\A_{int} \rightarrow \A_{\mathfrak{R}}$$ of the extensionalization function $h:\D \rightarrow  D_{-1} +\mathfrak{R}$.
\begin{definition} \textsc{Basic Intensional \textsc{FOL} Algebra:} \\ \label{def:intalgebra}
Intensional FOL algebra is a structure

$~\A_{int} =
~~(\D, Id, Truth, \{conj_{S}\}_{ S \in \P(\mathbb{N}^2)}, neg,
\{exists_{n}\}_{n \in \mathbb{N}})$, \\
$~~$ with binary operations  $~~conj_{S}:D_I\times D_I \rightarrow D_I$, unary operation  $~~neg:D_I\rightarrow D_I$, and unary operations $~~exists_{n}:D_{I}\rightarrow D_I$, such that for any extensionalization function $h \in \E$, and $u \in D_k, v \in D_j$, $k,j \geq 0$,

1. $~h(Id) = R_=~$ and $~h(Truth) = \{<>\}$, for $Id= I(\doteq(x,y))$ and $Truth = I(\top)$.

2. $~h(conj_{S}(u, v)) = h(u) \bowtie_{S}h(v)$, where $\bowtie_{S}$
is the natural join operation and $conj_{S}(u, v) \in D_m$ where $m = k + j- |S|$  if for every pair $(i_1,i_2) \in S$ it holds that $1\leq i_1 \leq k$, $1 \leq i_2 \leq j$ (otherwise $conj_{S}(u, v) \in D_{k+j}$).

3. $~h(neg(u))  = ~ \sim(h(u))=\D^k \backslash (h(u))$ (the complement of k-ary relation $h(u)$ in $\D^k$), if $k\geq 1$,
 where  $neg(u) \in D_k$. For $u_0 \in \D_0$, $~h(neg(u_0))= ~ \sim(h(u_0))  = \D^0 \backslash (h(u_0))$.

 4. $~h(exists_{n}(u)) = \pi_{-n}(h(u))$, where $\pi_{-n}$ is the projection operation which eliminates $n$-th column of a relation and $exists_n(u) \in D_{k-1}$ if $1
\leq n \leq k$  (otherwise $exists_n$ is the identity function).
\end{definition}
Notice that for $u,v \in D_0$, so that $h(u),h(v) \in \{f,t\}$,

$~h(neg(u))  = \D^0 \backslash (h(u)) = \{<>\}
\backslash (h(u)) \in \{f,t\}$, and

$h(conj_\emptyset(u,v)) = h(u)\bowtie_\emptyset h(v)  \in \{f,t\}$.\\
We define a derived operation $~~union:(\P(D_i)\backslash \emptyset)
\rightarrow D_i$, $i \geq 0$, such that, for any $B = \{u_1,...,u_n\} \in \P(D_i)$ and $S = \{(l,l)~|~1 \leq l \leq i \}$ we have that
\begin{equation} \label{eq:union}
 union(\{u_1,...,u_n\})=\left\{
    \begin{array}{ll}
    ~u_1, & \hbox{if $n= 1$}\\
      neg(conj_S(neg(u_1),conj_S(neg(u_2),...,neg(u_n))...), & \hbox{otherwise}
       \end{array}
  \right.
\end{equation}
Than we obtain that for $n \geq 2$:

$h(union (B)) = h(neg(conj_S(neg(u_1),conj_S(neg(u_2),...,neg(u_n))...)$

$ = \D^i\backslash((\D^i\backslash h(u_1))
\bowtie_{S}...\bowtie_{S}(\D^i\backslash h(u_n)))
= \D^i\backslash((\D^i\backslash h(u_1))
\bigcap...\bigcap(\D^i\backslash h(u_n)))$

$= \bigcup\{ h(u_j)~|~1 \leq j \leq n\}$, that is,
\begin{equation} \label{eq:unionA1}
h(union (B)) =  ~\bigcup \{h(u) ~|~u \in B\}
\end{equation}
Note that it is valid also for the propositions in $u_1,u_2 \in D_0$, so that $h(union(u_1,u_2)) = h(u_1) \bigcup h(n_2) \in \{f,t\}$ where $f$ is empty set $\emptyset$ while $t$ is a singleton set $\{<>\}$ with empty tuple $<>$, and hence the join $\{<>\}\bowtie \emptyset = \emptyset$ and $\{<>\}\bowtie \{<>\} = \{<>\}$.

Thus, we define the following homomorphic extension  $$I:\A_{FOL} \rightarrow \A_{int}$$ of the intensional
interpretation $I:\L \rightarrow \D$ for the  formulae in syntax algebra $\A_{FOL}$ from Definition \ref{coro:intensemant}:
\begin{enumerate}
  \item The logic formula $\phi(x_i,x_j,x_k,x_l,x_m) \wedge_S \psi
(x_l,y_i,x_j,y_j)$ will be intensionally interpreted by the concept
$u_1 \in D_7$, obtained by the algebraic expression $~
conj_{S}(u,v)$ where $u = I(\phi(x_i,x_j,x_k,x_l,x_m)) \in D_5, v =
I(\psi (x_l,y_i,x_j,y_j))\in D_4$ are the concepts of the virtual
predicates $\phi, \psi$, relatively, and $S = \{(4,1),(2,3)\}$.
Consequently, we have that for any two formulae $\phi,\psi \in \L$
and a particular  operator $conj_S$ uniquely determined by tuples of
free variables in these two formulae, $I(\phi \wedge_S \psi) =
conj_{S}(I(\phi),I(\psi))$.
  \item The logic formula $\neg \phi(x_i,x_j,x_k,x_l,x_m)$ will be
intensionally interpreted by the concept $u_1  \in D_5$, obtained by
the algebraic expression $~neg(u)$ where $u$ is the concept of the virtual predicate $\phi$, $u = I(\phi(x_i,x_j,x_k,x_l,x_m)) \in D_5$. Consequently, we have that for any formula $\phi \in \L$, $~I(\neg \phi) = neg(I(\phi))$.
  \item The logic formula $(\exists_3) \phi(x_i,x_j,x_k,x_l,x_m)$ will
be intensionally interpreted by the concept $u_1  \in D_4$, obtained
by the algebraic expression $~exists_{3}(u)$ where $u =
I(\phi(x_i,x_j,x_k,x_l,x_m)) \in D_5$ is the concept of the virtual
predicate $\phi$. Consequently, we have that for any formula $\phi
\in \L$ and a particular operator $exists_{n}$ uniquely determined
by the position of the  existentially quantified variable in the
tuple of free variables in $\phi$ (otherwise $n =0$ if this
quantified variable is not a free variable in $\phi$), $~I((\exists_n)\phi) = exists_{n}(I(\phi))$.\index{Frege/Russel semantics}
\end{enumerate}
So, we obtain the following two-steps interpretation of FOL based on two homomorphisms, intensional $I$, and extensional $h$:
 \begin{equation} \label{diag:FregRussle}
\begin{diagram}
   &&& &     \A_{int}~ (concepts/meaning) &&& &\\
  &&\ruTo^{intensional~interpret.~I} && \frac{Frege/Russell}{semantics}  &&\rdTo^{h ~(extensionalization)} &&\\
 \A_{FOL}~(syntax)  &&&&           &&&& \A_{\mathfrak{R}} ~(denotation)   \\
\end{diagram}
\end{equation}
We can enrich the expressivity of such a minimal FOL intensionality  by new modal operators, or in different way provided in what follows. As, for example, in Bealer's intensional FOL, where he introduced the intensional abstraction operator, which will be considered in rest of this section, as a significant enrichment of the intensional FOL considered above.

 In reflective languages, reification data is causally connected to the related \index{reification} reified aspect such that a modification to one of them affects the other. Therefore, the reification data is always a faithful representation of the related reified aspect. \emph{Reification data} is often said to be made a \emph{first class object}. In programming language design, a first-class citizen (also type, object, entity, or value) in a given programming language is an entity which supports all the operations generally available to other entities. These operations typically include being passed as an argument, returned from a function, modified, and assigned to a variable. The concept of first and second-class objects was introduced by Christopher Strachey in the 1960s when he contrasted real numbers (first-class) and procedures (second-class) in ALGOL.

 In FOL we have the variables as arguments inside the predicates, and terms which can be assigned to variables are first-class objects while the predicates are the second-class objects. When we transform a virtual predicate into a term, by using intensional abstraction operator, we transform a logic formula into the first class object to be used inside another predicates as first-class objects. Thus, abstracted terms in the intensional FOL are just such abstracted terms as reification of logic formulae. For example, the sentence "Marco thinks \emph{that Zoran runs}", expressed by $thinks(\emph{Marco}, \lessdot runs(\emph{Zoran})\gtrdot)$ by using binary predicate $thinks$ and unary predicate $runs$ where the ground atom $runs(\emph{Zoran})$ is reified into the predicate $thinks$.

 If $\phi(\textbf{x})$ is a formula (virtual predicate) with a list (a tuple) of free variables in $\textbf{x} =(x_1,...,x_n)$ (with ordering from-left-to-right of their appearance in $\phi$), and  $\alpha$ is its subset of \emph{distinct} variables,
 then $\lessdot \phi(\textbf{x}) \gtrdot_{\alpha}^{\beta}$ is a term, where $\beta$ is the remaining set of free variables  in $\textbf{x}$. The externally quantifiable variables are the \emph{free} variables not in $\alpha$. When $n =0,~ \lessdot \phi \gtrdot$ is a term which denotes a proposition, for $n \geq 1$ it denotes  a n-ary concept.
 \begin{definition} \label{def:abstrConv} \textsc{Intensional abstraction convention}:

 From the fact that we can use any permutation of the variables in a given virtual predicate,  we introduce the convention that
 \begin{equation}\label{eq:abstrctConv}
 \lessdot \phi(\textbf{x})\gtrdot_{\alpha}^{\beta}~~ is~a~ term~ obtained~ from~ virtual ~ predicate ~~\phi(\textbf{x})
 \end{equation}
 if $\alpha$ is \textsl{not empty}   such that  $\alpha\bigcup\beta$ is  the set of all variables in the list (tuple of variables)  $\textbf{x} = (x_1,...,x_n)$ of the virtual predicate (an open logic formula) $\phi$,  and $\alpha\bigcap\beta = \emptyset$, so that $|\alpha|+|\beta| = |\textbf{x}| = n$.
 Only the variables in $\beta$ (which are the only free variables of this term), can be quantified. If $\beta$ is empty then $\lessdot \phi(\textbf{x})\gtrdot_{\alpha}$ is a \emph{ground term}. If $\phi$ is a sentence and hence both $\alpha$ and $\beta$ are empty, we write simply $\lessdot \phi \gtrdot$ for this ground term.
 \end{definition}
 More about this general definition of abstract terms can be find in \cite{Majk22}. In this paper we will use the most simple cases of ground terms $\lessdot \phi \gtrdot$, where $\phi$ is a sentence.
\section{Four-levels Robot's Brain Structure \label{sec:HRobot}}
Let us consider a model of robot for understanding language about space and movement in realistic situations \cite{Koll10,Tell10}, as finding video clips that match a spatial language description such as "People walking through the kitchen and then going to the dining room" and following natural language commands such as "Go down the hall towards the fireplace in the living room."

Video retrieval is a compelling application: in the United States alone, there are
an estimated 35 million surveillance cameras installed, which record four billion hours of video per week. Analyzing and understanding the content of video
data remains a challenging problem. A spatial language interface to video data can
help people naturally and flexibly find what they are looking for in video collections. Studying language used to give directions could enable a robot to understand natural language directions. People talk to robots even if they do not have microphones installed, and it makes sense to build systems that understand what they say. A robot that understands natural language is easy for anyone to use without special training. By using the deductive properties of the IFOL, the robot can make logic deductions as well about the facts that it visually recognized and also to obtain its own autoepistemic deductions about obtained knowledge, as shortly explained in introduction, by using intensional abstractions in Definition \ref{def:abstrConv}.

Consequently, I will focus on a narrow subset of a natural language, grounding that language in data collected from a real world. This strategy has two benefits. First, it decreases the scope of the language understanding problem, making it more tractable. Second, by choosing a semantically deep core domain, it offers an opportunity to explore the connection between linguistic and non-linguistic concepts.

The linguistic structure extracted from spatial language expressions and many of
the features in the model for spatial relations are based on the theories of Jackendoff \cite{Jack83}, Landau and Jackendoff \cite{LaJa93} and Talmy \cite{Talm05}. For example, the implementation of the mining of "across" in \cite{Talm05} is obtained by an algorithm (of robot's AI neuro-system) for computing the axes a figure imposes on a ground, and set of features which quantify "roughly perpendicular", using a machine learning algorithm to fine-tune the distinctions by training on labeled data. Regier \cite{Regi92} built a system that assigns labels such as "through" to move showing a figure relative to a ground object. Bailey \cite{Bail97} developed a model for learning the meanings of verbs of manipulation such as "push" and "shove". Kelleher and Costello \cite{KeCo09}  built models for the meanings of static spatial prepositions such as "in front of" and "above". Siskind \cite{Sisk01} created a system for defining meanings for words such as "up" and "down."  The framework reasons about formal temporal relations between primitive force-dynamic properties such as "supports" and "touches" and uses changes in these properties to define meanings for verbs. His framework focuses on word-level event recognition and features, etc..

Reasoning about movement and space is a fundamental competence of humans and many animals. Humans use spatial language to tell stories and give directions, abstracting away the details of a complex event into a few words such as "across the kitchen." A system that understands spatial language could be directly useful to people by finding video that matches spatial language descriptions, or giving natural language directions. We will consider a robot which retrieves video clips that match a natural language description using a probabilistic graphical model that maps between natural language and paths in the environment \cite{Koll10}.

In this particular environment, spatial relations are modeled as probabilistic
distributions for recognizing words paired with scenes. The distributions are trained from labeled examples using a set of geometric features that capture the semantics of spatial prepositions. The distribution  modeled is the probability of a particular spatial relation given a trajectory  and an object in the environment. This distribution corresponds to the probability that a spatial relation such as "across" or "to" describes a particular trajectory and landmark. The input to the model is the geometry of the path and landmark object; the output is a probability that the spatial relation can be used to describe this scene.
These distributions are trained using labeled path examples, and in robot's brain correspond to its AI neuro-system. The system learns distributions for spatial relations, for example, by using a naive Bayes probabilistic
model.

So, now we can focus to the integration of such robot's AI neuro-system with its AI symbolic system based on three natural language cognitive levels: The \emph{syntax} of a particular natural language (French, English, etc..) its \emph{semantic logic structure} (transformation of parts of the language sentences into the logic predicates and definition of corresponding FOL formulae) and its corresponding \emph{conceptual structure}, which differently from the semantic layer that represents only the logic's semantics, represents the composed meaning of FOL formulae.

In this example, we focus on spatial
language search of people’s motion trajectories which are automatically extracted from video recorded by stationary overhead cameras. The system takes as input a natural language query, a database of surveillance video from a particular environment and the locations of non-moving objects in the environment. When the robot performs video retrieval by its AI neuro system, clips are
returned in order according to the joint probability of the
query and the clip. Thus, for each video clip in given database,
this robot's neuro system  computes the probability that considered clip satisfies a natural language query, parsed into logic FOL formula (second natural language semantic level) and consequently into intensional algebra $\A_{int}$ term with intensional concepts which labels are grounded by robot's neuro system processes (algorithms). Let $\N\L$ be a given natural language. If we denote the set of finite nonempty lists of a given natural language words by  $\N\L_{list}$, then this parsing can be represented by a \emph{partial} mapping
\begin{equation} \label{vp}
   pars:\N\L_{list}\rightarrow \L
\end{equation}
where $\L$ is the set of logic formulae of intensional FOL.

We suppose that the concepts in the conceptual structure expressed by the intensional algebra $\A_{int}$ of \emph{atomic} concepts $u \in \D$, and their corresponding logic atoms expressed by virtual predicates $\phi(\textbf{x}) \in \L$ of FOL are the part of  innate robot's knowledge, such that for robot's innate and unique intensional interpretation $I:\L\rightarrow \D$, $u = I(\phi(\textbf{x}))$.
Moreover, we suppose that robot has a parser capability to transform the sentences of particular natural language into the formulae of FOL with innate set of the atoms expressed by virtual predicates.

In this example we consider the predicates of IFOL as the verbs (V) of natural language, as follows $$Find(x_1,x_2,x_3,x_4)$$ where the time-variable $x_1$ (with values "in past", "in present", "in future") indicates the time of execution of this recognition-action, the variable $x_2$ is used for the subject who executes this action (robot in this case), the variable $x_3$ is used for the object given to be eventually recognized (in this case a video clip) and $x_4$ for the statement (users query) that has to be satisfied by this object, and virtual predicate $$Walk(x_1,x_2,x_3,x_4,x_5)$$ where the time-variable $x_1$ (with values "in past", "in present", "in future") indicates the time of execution of this action, variable $x_2$ for the figure (F) that moves ("person", "cat", etc..), $x_3$ for the initial position of walking figure (defined by the spatial relation (SR) "\emph{from}", for example "from the table") , $x_4$ for the intermediate positions during movement of the figure (defined by (SR) "\emph{through}", for example  "through the corridor"), and $x_5$ for the final position of figure (defined by (SR) "\emph{to}", for example "to the door").

The robot takes as input a natural language query, a database of surveillance video from a particular environment and the locations of non-moving objects in the environment. It parses the query into a semantic structure called a spatial description clause (SDC) \cite{Tell10}.  An SDC consists of a figure (F), a verb (V), a spatial relation (SR), and a landmark (L). The system extracts SDCs automatically using a conditional random field chunker. Let us consider the example illustrated in Figure 3 in \cite{Tell10} of a natural language query $nq\in \N\L_{list}$, defined by a sentence:
\begin{center}
"The person walked \emph{from} the couches in the room \emph{to} the dining room table"
\end{center}
which is composed by two SDC with the first one
\begin{enumerate}
  \item (F) = "the person"
  \item (V) = "walked"
  \item (SR) = "from"
  \item (L) = "the couches in the room"
\end{enumerate}
and the second SDC,
\begin{enumerate}
    \item (SR) = "to"
  \item (L) = "the dining room table"
\end{enumerate}
\textbf{Remark}: Note that all SDC components different from (V), are particulars in $D_{-1}$ in PRP domain $\D$, provided by Definition \ref{def:Idomain}. The sense (mining) of the components (F) and (L) are grounded by the machine-learning video-recognition processes of the robot, that is by its neuro systems. The sense of the (SR) components is grounded by the meaning of the spatial relations, provided by different authors methods, mentioned previously, and implemented by particular robots processes.\\
What we need in next is to extend this grounding also to the virtual predicates of the FOL open formulae in $\L$.
\\$\square$\\
Consequently, from these Spatial Description clauses, for the (V) of the past-time verb (V) "to walk", the semantic logic structure recognized by robot is the sentence $\phi = pars(nq) \in\L$,  obtained from (\ref{vp})
so that, based on the virtual predicate $toWalk$,  the sentence $\phi$ is
\begin{equation} \label{vp1}
Walk(in~past,person,from~ the~ couches~ in ~the~room,NULL,\\to~ the~ dining~ room ~table)
\end{equation}
Note that the inverse parsing of such logic sentence $\phi$ to natural language sentence is directly obtained, so that the robot can translate its semantic logic structures into natural language to communicate by voice to the people.

We consider that each  grammatically \emph{plural} word name "videoclips", robot can define by generalization by creating the virtual unary predicate $videoclips(y)$, such that its intensional \emph{concept} $ u_2= I(videoclips(y)) \in D_1$ in PRP domain, whose meaning is grounded by  robots patern-recognition  process fixed by a machine learning method. In a similar way, each unary concept of visual objects can be created by robot by a machine learning method for enough big set of this type of objects.

So, each  grammatically \emph{singular} word name, like "John's videoclip" is a particular (element of $D_{-1}$) in PRP domain, whose meaning is grounded by the internal robot's image of this particular videoclip, \emph{recognized} as such by robots patern-recognition process.
Thus, for a given extensionalization function $h$ in (\ref{eq:ExtenFunct}), and fixed robot's intensional mapping $I$, from the diagram (\ref{diag:FregRussle}), we obtain that the set $C$, of video clips in a given database of videoclips presented to this robot, is equal to
\begin{equation} \label{v1}
 C = h(I(videoclips(y)))
\end{equation}
Consequently, the human command in natural language $nc \in \N\L_{list}$ to this robot,
\begin{center}
"Find videoclip  such that $\phi$ in the given set of videoclips"
\end{center}
(where $\phi$ has to be substituted by the sentence above) is parsed by robot into its second level (semantic logic structure) by virtual predicate $Find$ of the verb "to find" (\emph{in present}) and a variable $y$ of type "videoclip" (objects of research) and substituting "that $\phi$" by abstracted term $\lessdot\phi\gtrdot$, and by substituting  "in the given set of" with the logic conjunction connective $\wedge_S$ of the IFOL expressed, from (\ref{vp}), by the following formula $\psi(y) = pars(nc)$
\begin{equation} \label{v2}
   Find(in present,me,y,\lessdot\phi\gtrdot)\wedge_S videoclips(y)
\end{equation}
where $S = (2,1)$ for joined variables in two virtual predicates.

The meaning of the unary concept $u_1 = I(Find(in~present,me,y,\lessdot\phi\gtrdot))$, corresponding to the natural language subexpression "Find (me),videoclip such that $\phi$" of the command above, is  represented by its AI neuro system process of probabilistic recognition of video clips \cite{Tell10} satisfying the natural language query $\phi$ (In fact, $u_2$ is just equal to the name of this process of probabilistic recognition).

However, during execution of this process, the robot is able also to \emph{deduce} the truth of the \emph{autoepistemic sentence}, for a given assignment of variables $g:\V\rightarrow \D$, with $g(x_1) = in~ present$ and $g(x_2) = me$,
\begin{equation} \label{v3}
Know(x_1, x_2, \lessdot Find(in present,me,y,\lessdot\phi\gtrdot)\gtrdot_y)/g
\end{equation}
of the virtual predicate $Know(x_1, x_2, x_3)$, where the time-variable $x_1$ (with values "in past", "in present", "in future") indicates the time of execution of this action, the variable $x_2$ is used for the subject of this knowledge and $x_3$ is used for an abstracted term expression this particular knowledge). Thus, by using deductive properties of the true sentences of FOL, this autoepistemic sentence about its state of selfknowledge, the robot would be able to comunicate to humans this sentence, traduces in natural language as
\begin{center}
"I (me) know that I am (me) finding  \emph{videoclip} such that $\phi$"
\end{center}
From the fact that robot defined the type of the variable $y$  to be "videoclip", by traduction of the FOL deduced formula above into the natural language, this variable will be traduced in natural language by "videoclip".
In the same way, during the execution of the human command above, expressed by the FOL formula $\psi(y)$ in (\ref{v2}), with composed concept $u_3 = I(\psi(y)) \in D_1$, that is, by using the homomorphic property of intensional interpretation $I$,
\begin{equation} \label{v4a}
   u_3 = u_1\bowtie_S u_2
\end{equation}
the robot can deduce also the true epistemic sentence,  for a given assignment of variables $g:\V\rightarrow \D$, with $g(x_1) = in~ present$ and $g(x_2) = me$,
\begin{equation} \label{v4}
Know(x_1, x_2, \lessdot Find(in ~present,me,y,\lessdot\phi\gtrdot)\wedge_S videoclips(y)\gtrdot_y)/g
\end{equation}
and hence the robot would be able to communicate to humans this sentence, traduces in natural language as
\begin{center}
"I (me) know that I am (me) finding \emph{videoclip} such that $\phi$ in the set of videoclips"
\end{center}
Note that the subset of videoclips extracted by robot from a given set of videoclips  $C = h(u_2)$ in (\ref{v1}), defines the current extensionalization function $h$, in the way that this subset is
\begin{equation} \label{v5}
E = h(u_3) = h(u_1)\bowtie_S h(u_2) = h(u_1)\bowtie_S C = h(u_1) \subseteq C
\end{equation}
Thus, for the grounding of spatial language for video search, the robot's internal knowledge structure is divided into four levels, in ordering: natural language, semantic logic structure,  conceptual structure and neuro structure, as represented by the following diagram (only two continuous arrows (intensional mapping $I:\L \rightarrow D_I$ where $D_I = D_0+D_1+...$ are the universals in PRP domain theory) represent the total mappings, while other (dots) are partial mappings)
 
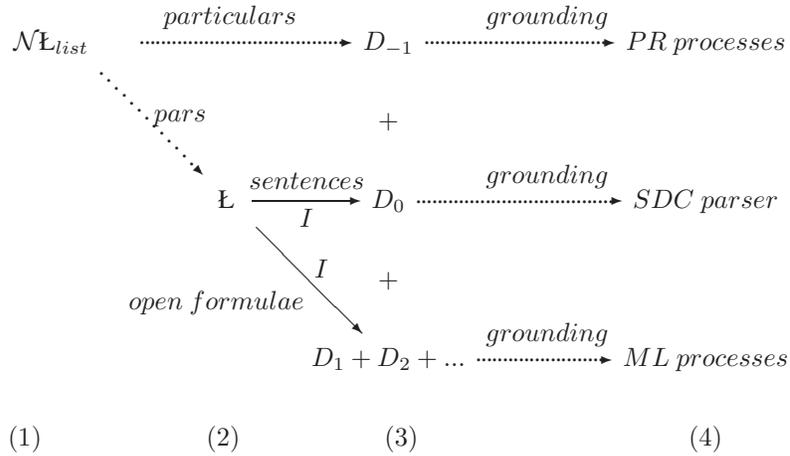
\begin{equation} \label{diag:FregRobKnowledge}
\begin{diagram}
   \N\L_{list}~~~~~~&&\rDotsto^{particulars} && D_{-1} &&\rDotsto^{grounding}&& PR ~processes\\
  &\rdDotsto^{pars}&            && + && &&\\
 &&\L~      &\rTo^{sentences}_I&          D_0 && \rDotsto^{grounding}&& SDC~parser    \\
 &&&\rdTo_{open~ formulae}^I     &          + && &&     \\
 &&&    &          D_1+D_2+... && \rDotsto^{grounding}&& ML~processes    \\
 ~~(1)~~~~~~~~~~~~~~~~&&~~~~ (2)~~~~~~&&~~~~       (3) && && (4)    \\
 Nat.Lang.~~~~~~~~~~~~~~~~&&~~~~ Log.semantic~ sys.~~~~~~~~&&~~~~~~~~~~ ~~         Conceptual ~sys. && && Neuro ~sys.    \\
\end{diagram}
\end{equation}
It is easy to see that the conceptual system, based on PRP domain $\D$ composed by particulars in $D_{-1}$ and universals (concepts) in $D_I = D_0+D_1+D_2+...$ of the IFOL, is the level of grounding of the natural language of the robot to its neuro system composed by the following processes:
\begin{enumerate}
  \item PR (Pattern Recognition) processes of recognition of the particulars. For example, for SDC components (F) "the person",  (L) "the couches in the  room" and "the dining room table", etc..
  \item SDC (Spatial Description Clauses) parser used for the sentences,  for example, for a natural language query $nq \in \N\L_{list}$ that is, logical proposition (sentece) $\phi = pars(nq) \in \L$ in (\ref{vp1}), which is labeled by its intensional proposition label $I(\phi) \in D_0$. Thus, the grounding of $nq$ is obtained by linking its intensional proposition $I(pars(nq))$ in PRP to the SDC parser process (part of robot's neuro system).
  \item ML (Machine Learning) processes, like that used for the recognition of different types of classes (like the set of videoclips). For example, for the language plural world "videoclips" in $\N\L_{list}$, such that $pars("videoclips") = videoclips(y) \in \L$ with its intensional unary concept $u_2 = I(videoclips(y)) \in D_1$ which is grounded to robot's ML process for the "videoclips".
\end{enumerate}
Note that, while the top line in the diagram (\ref{diag:FregRobKnowledge}) is the ordinary component of the natuaral language grounding developed by robot's neuro system, the two lines bellow is the new robots knowledge structure of the added \emph{symbolic AI system} based on the Intensional First Order Logic and its grounding to robot's processes (its neuro AI system), by which the robot is able to provide logic deductive operations  and autoepistemic self-reasoning about its current knowledge states and communicate it to humans by using natural languages.
\section{A Short Introduction to Robots Autoepistemic Deduction}
In my recent book  it has been demonstrated that the Intensional FOL  \cite{Majk22} has a conservative Tarski's semantics, shown also in this paper (only partially) by Definition \ref{eq:abstrctConv}, with interpretations (see the diagram in (\ref{diag:FregRussle}))
$$I_T^* = h\circ I :\A_{FOL} \rightarrow \A_{\mathfrak{R}}$$
as the ordinary (extensional) FOL with well known its deductive properties.

By introduction  of the abstraction operators with autoepistemic capacities, expressed by the $Know$ predicate in previous section, we do not use more a \emph{pure} logical deduction of the standard FOL, but a kind of autoepistemic deduction \cite{Majk04ph,MajkA04} with a proper set of new axioms. However, the autoepistemic logic is introduced as a propositional logic \cite{MaTr91} with added universal modal operator, usually written $K$,  and the axioms:
\begin{enumerate}
  \item Reflexive axiom \textbf{T}:   $~~K\phi \Rightarrow \phi$
  \item Positive introspection axiom \textbf{4}:  $~~K\phi\Rightarrow KK\phi$
  \item Distributive axiom \textbf{K}:   $~~(K\phi \wedge K(\phi\Rightarrow\psi)) \Rightarrow K\psi$
\end{enumerate}
for any proposition formulae $\phi$ and $\psi$,
while $Know$ in IFOL is a \emph{predicate} and not modal (Kripke-like) operator $K$.

It has been demonstrated that intensional enrichment of the standard (extensional) FOL, provided by Definition 14 in \cite{Majk22}, is a kind of modal predicate logic FOL$_{\K}(\Gamma)$, where the set of explicit possible world $\W_e$ is equal to the set $\mathfrak{I}_T(\Gamma)\}$ of Tarski's interpretations  $I_T^* = h\circ I$ (this composition is provided by diagram in (\ref{diag:FregRussle})) of the standard FOL with a given set of assumptions $\Gamma$, that is, for a prefixed intensional interpretation of robot, this set of possible worlds is equal to the set of the extensionalization functions $h \in \E$ of robot's IFOL. It has been demonstrated that in such a minimal intensional enrichment of standard (extensional) FOL, we obtain exactly the Montague's definition of the intension (see Proposition 5 in \cite{Majk22}).

We recall that each robot's extensionalitation function $h\in \E$ in (\ref{eq:ExtenFunct}) is indexed by the time-instance. The actual robot's world extensionalization function (in the current instance of time) is denoted by $\hbar$, and determines the current robot's knowledge.
 Clearly, the robots knowledge changes in time and hence determines the extensionalization function $h \in \E$ in any given instance of time, based on robots experiences. Thus, as for humans, also the robot's knowledge and logic is a kind of temporal logic, and evolves with time.

 Note that the explicit (conscious) robot's  knowledge in actual world $\hbar$ (current time-instance) here is represented by the ground atoms of the $Know$ predicate, for a given assignments $g:\V\rightarrow \D$, that is, $g\in \D^\V$,
 \begin{equation} \label{eq:esem2}
  Know(y_1,y_2,\lessdot \psi(\textbf{x})\gtrdot^\beta_\alpha)/g = Know(g^*(y_1),g^*(y_2), g^*(\lessdot \psi(\textbf{x})\gtrdot^\beta_\alpha))
  \end{equation}
  with  $\{y_1,y_2\}\bigcup \beta \bigcup \alpha \subseteq \V$,  such that $g^*(y_1) = in ~ present$ and $g^*(y_2) = me$ (the robot itself),  for the extended assignments $g^*:\T \rightarrow \D$,  where the set of terms $\T$ of IFOL is composed by the set $\V$ of all variables used in the defined set of predicates of robot/s IFOL, by the set of FOL constants and by the set of  \emph{abstracted terms} in (\ref{eq:abstrctConv}), such that (from Definition 17 in \cite{Majk22}):
  \begin{enumerate}
  \item $g^*(t) = g(x) \in \D$ if the term $t$ is a variable $x \in\V$.
  \item $g^*(t)  \in \D$ is the Tarski's interpretation  of the FOL constant (nullary function) if the term $t$ is a constant $c$.
  \item If $t$ is an abstracted term obtained for a formula $\phi$, $\lessdot \phi(\textbf{x}) \gtrdot_{\alpha}^{\beta}$,  then
\begin{equation} \label{eq:assAbTerm}
  g^*(\lessdot \phi(\textbf{x})\gtrdot_{\alpha}^{\beta}) =_{def}
    \left\{
    \begin{array}{ll}
   I(\phi(\textbf{x}))~~ \in D_{|\alpha|}, & \hbox{if  $\beta$ is  empty}\\
       I(\phi[\beta/
g(\beta)])~~ \in D_{|\alpha|}, & \hbox{otherwise}
       \end{array}
  \right.
  \end{equation}
where $g(\beta) = g(\{y_1,..,y_m\}) = \{g(y_1),...,g(y_m)\}$ and $[\beta
/g(\beta)]$ is a uniform replacement of each i-th variable in the
set $\beta$ with the i-th constant in the set $g(\beta)$. Notice that $\alpha$ is the set of all free variables in the formula $\phi[\beta /g(\beta)]$.
\end{enumerate}
so that in the actual world $\hbar$, the known fact (\ref{eq:esem2}) for robot becomes the ground atom
 \begin{equation} \label{eq:esem3}
  Know(y_1,y_2,\lessdot \psi(\textbf{x})\gtrdot^\beta_\alpha)/g = Know(in ~ present,me, I(\psi[\beta/g(\beta)]))
  \end{equation}
  which is true in actual word, that is, from proposition (intensional concept)\\ $u=I(Know(in ~ present,me, I(\psi[\beta/g(\beta)])))\in D_0$, we obtain \\$\hbar(u) = \hbar(I(Know(in ~ present,me, I(\psi[\beta/g(\beta)])))) = t$.

  \textbf{Remark}: Note that for the assignments $g:\V\rightarrow \D$, such that $g(y_1)= in ~future$ and $g(y_2)$ we consider robot's hypothetical knowledge in future, while in the cases when $g(y_1) = in ~past$ we consider what was robot's knowledge in the past. Consequently, generally the predicates of IFOL for robots, based on the dynamic changes of its knowledge has to be indexed by the time-instances (which are possible worlds of IFOL), for example by using an additional predicate's variable for them.  In the examples in next, we will consider only the case of robot's current knowledge (in the actual world with extensional function $\hbar$) when $g(y_1) =in ~present$, so we avoid the introduction of the time-instance variable for the predicates; only at the remark at the end of this section we will show how to use time-variable $\tau$.
\\$\square$\\
  From the fact that we do not use the modal Kripke universal modal operator $K$, the principle of necessitation rule $\textbf{N}$ for modal logics, which for a given proposition (sentence) $\phi$ derives the knowledge fact $K\phi$, here in predicate based IFOL, the robots current knowledge (ground atoms of predicate $Know$) is directly derived from its experiences (based on its neuro-system processes that robot is using in this actual world), in an analog way as human brain does:
  \begin{itemize}
    \item As an activation (under robot's attention) of its neuro-system process, as a consequence of some human command to execute some particular job.
    \item As an activation of some process under current attention of robot, which is part of some complex plan of robot's activities connected with its general objectives and services.
  \end{itemize}
  In both cases, for a given assignment $g:\V\rightarrow \D$ of virtual predicate $\phi(\textbf{x})[\beta/g(\beta)]$ with the \emph{set} of variables  $\overline{\textbf{x}} =\beta \bigcup \alpha$, which  \emph{concept} $I(\phi(\textbf{x})[\beta/g(\beta)]) \in D_{|\alpha|}$ is grounded by this particular process, is transformed into abstracted term and hence robot's knowledge system generates the new ground knowledge atom $Know(y_1,y_2,\lessdot \phi(\textbf{x})\gtrdot^\beta_\alpha)/g$ with $g(y_1) = in ~presence$ and $g(y_2) = me$, in robot's temporary memory.\\
   \textbf{Remark}:  We consider that only robot's experiences (under robot's attention) are transformed into the ground atoms of the $Know$ predicate, and the required (by robot) deductions from them (by using general FOL deduction extended by the three epistemic axioms) are transformed into ground atoms of $Know$ predicate, and hence are saved in robot's temporary memory as a part of robot's \emph{conscience}.\\ Some background process (unconscious for the robot) would successively transform these temporary memory knowledge into permanent robot's knowledge in an analog way as it happen for humans.
   \\$\square$\\
  Thus, the three epistemic axioms of epistemic modal logic with modal operator $K$,  used to obtain deductive knowledge, can be traduced in IFOL by the following axioms for the predicate $Know$, which in fact define the semantics of this particular $Know$ predicate, as follows:
\begin{enumerate}
  \item   The modal axiom $\textbf{T}$,  in IFOL is represented by the axiom, for each abstracted term $\lessdot \psi(\textbf{x})\gtrdot^\beta_\alpha$ and assignment $g:\{y_1,y_2\}\bigcup\beta \rightarrow \D$,
\begin{description}
\item[(a)] If $\alpha$ is empty
\begin{equation} \label{eq:asio1}
  Know(y_1,y_2,\lessdot \psi(\textbf{x})\gtrdot^\beta)/g \Rightarrow \psi[\beta/g(\beta)]
  \end{equation}
  \item[(b)] If $|\alpha|\geq 1$ and  for the intensional concept $u_1 = I(\psi[\beta/g(\beta)]) \in D_{|\alpha|}$,\\
  $\hbar(u_1) = \{g_i(\textbf{y})\mid g_i \in \D^\alpha, 1\leq i\leq n$, and $\hbar(I(\psi[\alpha/g_i(\alpha)][\beta/g(\beta)])) =t \}$\\ with $g_i \neq g_j$ if $i\neq j$ and  the tuple of hidden variables $\textbf{y}$ in the virtual predicate $\psi[\beta/g(\beta)]$,
  \begin{equation} \label{eq:asio1f}
  Know(y_1,y_2,\lessdot \psi(\textbf{x})\gtrdot^\beta_\alpha)/g \Rightarrow
  Know(y_1,y_2,\lessdot \psi[\alpha/g_1(\alpha)]\wedge...\wedge \psi[\alpha/g_n(\alpha)]\gtrdot^\beta)/g
  \end{equation}
  This axiom shows how the robot's experience of execution of the process (described by abstracted term $\lessdot \psi(\textbf{x})[\beta/g(\beta)]\gtrdot_\alpha$), to which the intensional concept $u_1$ is grounded, transforms the true facts obtained by robot's neuro-system (of this process, which results are still the parts of robots unconscious knowledge) into the symbolic-AI FOL formula $$(\psi[\alpha/g_1(\alpha)]\wedge...\wedge \psi[\alpha/g_n(\alpha)])[\beta/g(\beta)]$$
  So by using axiom (\ref{eq:asio1}), and FOL deduction, these deductive properties of the robot can deduce  any true single fact (logical sentence) $\psi[\alpha/g_i(\alpha)][\beta/g(\beta)]$ derived by its neuro-system process, and to render it to robot's consciousness as a single  known fact $Know(y_1,y_2,\lessdot \psi[\alpha/g_i(\alpha)]\gtrdot^\beta)/g$.
  \end{description}
   In the case \textbf{(a)}, when $\alpha$ is empty, from (\ref{eq:assAbTerm}) with
    \begin{equation} \label{eq:asio2}
  u_1 =  g^*(\lessdot \psi(\textbf{x})\gtrdot^\beta)= I(\psi[\beta/g(\beta)]) \in D_0
  \end{equation}
such that $\hbar(u_1) = \hbar(I(\psi[\beta/g(\beta)]))=t$, that is, the sentence $\psi[\beta/g(\beta)]$ is true, so from the fact that  the left side ground atom of axiom's implication in (\ref{eq:asio1})  is equal to $Know(g(y_1),g(y_2), u_1)$,  this $\textbf{T}$  axiom (\ref{eq:asio1}) becomes
\begin{equation} \label{eq:asio1b}
  Know(g(y_1),g(y_2),u_1) \Rightarrow \psi[\beta/g(\beta)]
    \end{equation}
Note that the meaning of the intensional concept $u_1$ of the robot is grounded on robot's neuro-system process, which is just robot's current internal experience of what is he doing, and just because of that the robot's knowledge $Know(g(y_1),g(y_2),u_1)$ is true for him. So, this is really an reflexive axiom.\\
Consequently, the application of the \textbf{T} axiom \textbf{(a)}, allows the extraction from robot's conscious knowledge the \emph{logical sentences} which, successively, can be elaborated by robot's implemented deductive property of FOL in two ways:\\

a.1. To respond to some human natural language questions (parsed into a logical formula) and to verify if the response  is "yes" or "no", or "I do not know" (if robot's conscious knowledge is incomplete for such a question);\\

a.2. To deduce another sentences which then can be inserted in robot's  conscious knowledge as ground atoms of the predicate $Know$ (where this deduced sentence is represented as an abstracted term). This process (in background, or when robot is free of other concrete activities) can be considered as a kind of consolidation and completion of robot's knowledge based on previous experiences, in an analog way as it is done by human mind when we sleep.\\
\item The positive introspection axiom \textbf{4}:
\begin{equation} \label{eq:asio1b4}
 Know(y_1,y_2,\lessdot \psi(\textbf{x})\gtrdot^\beta_\alpha)/g \Rightarrow Know(g(y_1),g(y_2),\lessdot Know(y_1,y_2,\lessdot \psi(\textbf{x})\gtrdot^\beta_\alpha)/g)
 \end{equation}
 that is.
 \begin{multline} \label{eq:asio1b4a}
 Know(g(y_1),g(y_2),g^*(\lessdot \psi[\beta/g(\beta)]\gtrdot_\alpha)) \Rightarrow\\ Know(g(y_1),g(y_2),\lessdot Know(g(y_1),g(y_2),g^*(\lessdot \psi[\beta/g(\beta)]\gtrdot_\alpha)))
  \end{multline}
 which, in the case when $g(y_1) = in ~present$ and $g(y_2) = me$, is traduced in natural language by robot as:\\
 "I know that $\psi[\beta/g(\beta)]$" implies "I know that I know that $\psi[\beta/g(\beta)]$"\\
 where in the logic virtual predicate $\psi[\beta/g(\beta)]$ there are the hidden variables in $\alpha$, with extension $\hbar(u)$ of its intensional concept $u = g^*(\lessdot \psi[\beta/g(\beta)]\gtrdot_\alpha) = I(\psi[\beta/g(\beta)]) \in D_{|\alpha|}$.\\
  \item The distributive axiom \textbf{K} ("modal Modus Ponens"):
  \begin{multline} \label{eq:asio1b4}
 (Know(y_1,y_2,\lessdot \psi(\textbf{x})\gtrdot^\beta_\alpha)/g \wedge Know(y_1,y_2,\lessdot \psi(\textbf{x}) \Rightarrow \phi(\textbf{z})\gtrdot^{\beta\bigcup\beta_1}_{\alpha\bigcup\alpha_1})/g) \\
 \Rightarrow Know(y_1,y_2,\lessdot \phi(\textbf{z})\gtrdot^{\beta_1}_{\alpha_1})/g
    \end{multline}
    with the \emph{sets} of variables  $\alpha \bigcup \beta = \overline{\textbf{x}}$ and $\alpha_1 \bigcup \beta_1 = \overline{\textbf{z}}$. Or, equivalently,
  \begin{multline} \label{eq:asio1b4c}
 (Know(g(y_1),g(y_2),g^*(\lessdot \psi[\beta/g(\beta)]\gtrdot_\alpha)) ~\wedge\\ Know(g(y_1),g(y_2),g^*(\lessdot \psi[\beta/g(\beta)] \Rightarrow \phi[\beta_1/g(\beta_1)]\gtrdot_{\alpha\bigcup\alpha_1}))) \\
 \Rightarrow Know(g(y_1),g(y_2),g^*(\lessdot \phi[\beta_1/g(\beta_1)]\gtrdot_{\alpha_1}))
    \end{multline}
 Note that this axiom, when $\alpha$ and $\alpha_1$ are empty, is a way how the robot provides the conscious implications $\psi[\beta/g(\beta)] \Rightarrow \phi[\beta_1/g(\beta_1)]$,  which can be interpreted just as \emph{a rule} "if $\psi[\beta/g(\beta)]$ then $\phi[\beta_1/g(\beta_1)$", independently if they are its innate implemented rules (introduced in robot's knowledge when it is created) or if they are learned by robot's own experience.  In fact, this implication can be used only when it is true that some actual robot's experience which produced in its consciousness the knowledge $(Know(y_1,y_2,\lessdot \psi(\textbf{x})\gtrdot^\beta_\alpha)/g$ so that, from \textbf{T} axiom \textbf{(a)}, the sentence $\psi[\beta/g(\beta)]$ is true, as it is necessary for execution of the rule "if $\psi[\beta/g(\beta)]$ then $\phi[\beta_1/g(\beta_1)$", and hence to derive the true fact $\phi[\beta_1/g(\beta_1)$.
\end{enumerate}
Despite the best efforts over the last years, deep learning is still easily fooled \cite{NYCl15}, that is, it remains very hard to make any guarantees about how the system will behave given data that departs from the training set statistics. Moreover, because deep learning does not learn causality, or generative models of hidden causes, it remains \emph{reactive}, bound by the data it was given to explore \cite{Marc18}.

In contrast, brains act proactively and are partially driven by endogenous curiosity, that is, an internal, epistemic, consistency, and knowledge-gain-oriented drive. We learn from our actively gathered sensorimotor experiences and form conceptual, loosely hierarchically structured, compositional generative predictive models.
By proposed four-level cognitive robot's structure (\ref{diag:FregRobKnowledge}), IFOL allow robots to reflect on, reason about, anticipate, or simply imagine scenes, situations, and developments within in a highly flexible, compositional, that is, semantically meaningful manner. As a result, IFOL enables the robots to actively infer highly flexible and adaptive goal-directed behavior under varying circumstances \cite{Russ20}.

We are able to incorporate the emotional structure to robots as well, by a number of fuzzy-emotional partial mappings
\begin{equation} \label{eq:emotions}
E_i:\D \rightarrow [0,1]
\end{equation}
 of robots PRP intensional concepts,  for each kind of emotions $i\geq 1$: love, beauty, fear, etc. It was demonstrated \cite{Majk22} that IFOL is able to include any kind of many-valued, probabilistic and fuzzy logics as well.\\\\
\textbf{Example}: Let us consider the example provided in previous Section \ref{sec:HRobot} and how  robot, which is conscious of the fact that works  to respond to user question $\phi$ in (\ref{vp1}), and hence this robot knows that he initiated the process of recognition expressed by knowledge fact (\ref{v4}), for the assignment of variables $g:\{x_1,x_2\}\rightarrow \D$ with $g(x_1) = in~ present$ and $g(x_2) = me$,
\begin{equation}\label{fp5}
Know(x_1,x_2, \lessdot \psi(y)\gtrdot_y)/g
\end{equation}
where the logic formula $\psi(y)$ with a free variable $y$ is given by (\ref{v2}), i.e., is equal to conjunction $Find(in ~present,me,y,\lessdot\phi\gtrdot)\wedge_S videoclips(y)$, so that (\ref{fp5}) is equal to
\begin{equation}\label{fp5d}
Know(in ~present,me, \lessdot \psi(y)\gtrdot_y)
\end{equation}
How the robot becomes conscious of which video clips it recognized from a given set of videoclips (represented by the extension $C= h(I(videoclips(y))) = h(u_2)$ of the predicate $videoclips(y)$) we will show in next. In order to obtain this knowledge from the known fact (\ref{fp5}), the robot's "mind" can activate the internal neuro-process of the FOL-deduction, and hence to take the output  of these deducted facts from (\ref{fp5}) as new conscious knowledge (new generated ground atoms of its $Know$ predicate), as follows:
\begin{enumerate}
  \item From (\ref{v4a}), we have that unary PRP concept $u_3 = I(\psi(y)) \in D_1$  has the extension $E=h(u_3)$ given by (\ref{v5}),    composed by $n = |E| \geq 1$ elements, and hence from the \textbf{T} axiom \textbf{(b)} with $\alpha = \{y\}$,  we obtain that
      \begin{equation}\label{fp6}
\hbar(u_3) = \{g_i(y)\mid g_i \in \D^\alpha, 1\leq i\leq n, ~and~ \hbar(I(\psi[\alpha/g_i(\alpha)])) =t \}
\end{equation}
and hence this T axiom (\ref{eq:asio1f}), from (\ref{fp5d}), reduces to the logic implication
\begin{multline} \label{eq:asio1f2}
  Know(in ~present,me,\lessdot \psi(y)\gtrdot_y) \Rightarrow\\
  Know(in ~present,me,\lessdot \psi[y/g_1(y)]\wedge...\wedge \psi[y/g_n(y)]\gtrdot)
  \end{multline}
  \item So, from the true atom (\ref{fp5d}) and implication (\ref{eq:asio1f2}) by Modus Ponens rule of FOL we deduce the  formula (right side of implication (\ref{eq:asio1f2})),
     \begin{equation} \label{eq:asio1f3}
    Know(in ~present,me,\lessdot \psi[y/g_1(y)]\wedge...\wedge \psi[y/g_n(y)]\gtrdot)
  \end{equation}
  \item Now the deductive process can use the \textbf{T} axiom \textbf{(a)}, where $\beta$ is empty, and from it and (\ref{eq:asio1f3}), by Modus Ponens, deduce the conjunctive formula
      \begin{equation} \label{eq:asio1f4}
     \psi[y/g_1(y)]\wedge...\wedge \psi[y/g_n(y)]
  \end{equation}
  and hence, from this conjunctive formula by using FOL-deduction, to deduce that each ground atom $\psi[y/g_i(y)]$, for $1\leq i\leq n$, from (\ref{v2}) is equal to true fact
  \begin{equation} \label{eq:asio1f5}
  Find(in ~present,me,g_i(y),\lessdot\phi\gtrdot)\wedge videoclips(g_i(y))
  \end{equation}
  that is, to the true fact that this robot verified that the video clip $g_i(y)$ satisfies the users requirement $\phi$.
  \item The last step of this deduction process is to render these outputs of deduction conscious to this robot, that is, to transform the set of outputs in (\ref{eq:asio1f5}) into the set of \emph{known facts}, for $1\leq i\leq n$,
   \begin{equation} \label{eq:asio1f6}
  Know(in ~present,me,\lessdot Find(in ~present,me,g_i(y),\lessdot\phi\gtrdot)\wedge videoclips(g_i(y)\gtrdot)
  \end{equation}
\end{enumerate}
\textbf{Remark}: Note that the obtained robot's knowledge of the set in (\ref{eq:asio1f6}), from the known fact (\ref{fp5d}),  at the end of deduction is in robot's temporary memory. In order to render it permanent (by cyclic process of transformation of the temporary into permanent robot's memory), we need to add to any predicate of the robot's FOL syntax, also the time-variable as, for example, the first variable of each predicate (different from $Know$), instantiated in the known facts by the  \emph{tamestamp} value $\tau$ (date/time) when this knowledge of robot is transferred into permanent memory, so that the known facts  (\ref{eq:asio1f6}) in permanent memory would become
 \begin{equation} \label{eq:asio1f7}
  Know(in ~present,me,\lessdot Find(\tau, in ~past,me,g_i(y),\lessdot\phi\gtrdot)\wedge videoclips(g_i(y))\gtrdot)
  \end{equation}
 where the second value of the predicate $Find$, from \emph{in ~present} is modified into the value $in ~past$, and hence the FOL predicate $Find$ would be translated into natural language by the past time "have found" of this verb. So, the logic atom (\ref{eq:asio1f7}) can be  translated by robot into the following natural language autoepistemic sentence:\\
 "I know that I have found at $\tau$ the videoclip $g_i(y)$ which satisfied user requirement $\phi$."\\
 Note that, by  using the \textbf{T} axiom \textbf{(a)},  the robot can deduce, from permanent memory fact (\ref{eq:asio1f7}), the logic atom $$Find(\tau, in ~past,me,g_i(y),\lessdot\phi\gtrdot)\wedge videoclips(g_i(y))$$ as an answer to user question  and to respond in natural language simply by the sentence:\\
 "I have found at $\tau$ the videoclip $g_i(y)$ which satisfied user requirement $\phi$."
 \\$\square$\\
 This temporization of all predicates used in robot's knowledge is useful for robot to search all known facts in its permanent memory that are inside some time-interval as well.

 It can be used not only to answer directly to some human questions about robot's knowledge, but also to extract only a part of robot's knowledge from its permanent memory in order to be used for robot's deduction, and hence to answer to human more complex questions that require deduction of new facts not already deposited in explicit robot's known facts.

 \textbf{Remark}: We recall that this method of application of autoepistemic deduction  (for concepts such as \emph{knowledge}) can be applied to all other modal logic operators (for concepts such as  \emph{belief}, \emph{obligation}, \emph{causation}, \emph{hopes}, \emph{desires}, etc., for example by using \emph{deontic} modal logic that same statement have to represent a \emph{moral obligation} for robots), by introducing special predicates  for them with the proper set of axioms for their active semantics (fixing their meaning and deductive usage).

  By such fixing by humans of robot's unconciseness part with active semantics (which can not be modified by robots and their live experience) of all significant for human robot's concepts and their properties, we will obtain ethically confident and socially safe and non danger robots (controlled by public human  ethical security organizations for the production of robots with general strong-AI capabilities).
 \\$\square$
\section{Conclusions and future work}
Computation is defined purely formally or syntactically, whereas minds have actual mental or semantic contents, and we cannot get from syntactical to the semantic just by having the syntactical operations and nothing else… Machine learning is a sub-field of artificial intelligence. Classical (non-deep) machine learning models require more human intervention to segment data into categories (i.e. through feature learning). Deep learning is also a sub-field of machine learning, which attempts to imitate the interconnectedness of the human brain using neural networks. Its artificial neural networks are made up layers of models, which identify patterns within a given dataset. Deep learning can handle complex problems well, like speech recognition, pattern recognition, image recognition, contextual recommendations, fact checking, etc..

However, with this integrated four-level robot's knowledge system presented in diagram (\ref{diag:FregRobKnowledge}), where the last level represents the robot's neuro system containing the deep learning as well, we obtain that also the semantic theory of robot's intensional FOL is a procedural one, according to which sense is an abstract, pre-linguistic procedure detailing what operations to apply to what procedural constituents to arrive at the product (if any) of the procedure.

Weak AI, also known as narrow AI, focuses on performing a specific task, such as answering questions based on user input or playing chess. It can perform one type of task, but not both, whereas Strong AI can perform a variety of functions, eventually teaching itself to solve for new problems. Weak AI relies on human interference to define the parameters of its learning algorithms and to provide the relevant training data to ensure accuracy.

Strong AI (also known as full \emph{general AI}) aims to create intelligent robots that are quasi indistinguishable from the human mind. But just like a child, the AI machine would have to learn through input and experiences, constantly progressing and advancing its abilities over time. If researchers are able to develop Strong AI, the robot would require an intelligence more close to human's intelligence; it would have a self-aware consciousness that has the ability to solve problems, learn, and plan for the future.

However, since humans cannot even properly define what intelligence is, it is very difficult to give a clear criterion as to what would count as a success in the development of strong artificial intelligence.  Thus, we argue that this example, used for the spatial natural sublanguage, can be extended in a similar way  to cover  more completely the rest of human natural language, and hence the method provided by this paper is a main theoretical and philosophical contribution to resolve the open problem of how we can implement the deductive power based on IFOL for new models of robots heaving strong AI capacities.  Intensional FOL is able to represent the Intentional States (mental states such as beliefs, hopes, and desires), typical for human minds:

"\emph{Intentionality\footnote{A German philosopher and psychologist Franz Brentano (1838-1917) is best known for his reintroduction of the concept of intentionality, a concept derived from scholastic philosophy, to contemporary philosophy in his lectures and in his work Psychologie vom empirischen Standpunkt \cite{Bren874} (Psychology from an Empirical Standpoint). Brentano used the expression "intentional inexistence" to indicate the status of the objects of thought in the mind. Intentionality, based on the work of Austrian philosopher Alexius Meinong (1853-1920) a pupil of Franz Brentano, a realist known for his unique ontology, is the power of minds to be about something: to represent or to stand for things, properties and states of affairs. Intentionality is primarily ascribed to mental states, like perceptions, beliefs or desires, which is why it has been regarded as the characteristic mark of the mental by many philosophers. A central issue for theories of intentionality has been the problem of intentional inexistence: to determine the ontological status of the entities which are the objects of intentional states.\\
Meinong adopted the threefold phenomenological analysis of mental states that includes a mental act, its content and object of intention. Meinong wrote two early essays on David Hume, the first dealing with his theory of abstraction, the second with his theory of relations, and was relatively strongly influenced by British empiricism. He is most noted, however, for his edited book Theory of Objects (full title: Investigations in Theory of Objects and Psychology \cite{Mein904}), which grew out of his work on intentionality and his belief in the possibility of intending nonexistent objects. Whatever can be the target of a mental act, Meinong calls an "object."} is the fascinating property certain cognitive states and
events have in virtue of being directed, or about, something. When
ever we think, we think about something; whenever we believe, there is
something we believe; whenever we dream, there is something we dream
about. This is true of every episode of such diverse psychological phenomena as learning, imagining, desiring, admiring, searching for, discovering,
and remembering..}\\
\emph{Sometimes, they are directed towards logically more complex objects,
for instance, when we entertain a proposition, or fear a certain state of
affairs, or contemplate a certain depressing situation. But all of these
phenomena are to be contrasted with physical sensations, undirected
feelings of joy, sadness, depression, or anxiety, and with episodes of pain
and discomfort.}" pp. 10, \cite{Zalt88}\\

So, we are able to support such robot's physical sensations by using, for example, the manyvalued fuzzy logic values assigned to robot's PRP intensional "emotional" concepts, by mappings (\ref{eq:emotions}), representing  the feelings of joy, sadness, depression, or anxiety, etc.. as well. Hence, by using manyvalued logics embedded into IFOL, as explained in \cite{Majk22}, and autoepistemic deductive capacities provided in previous Section, the robots would be able to reason about their own  sensations and to communicate with humans.

Moreover, I argue that AI research should set a stronger focus on learning compositional generative predictive models (CGPMs), from robot's self-generated sensorimotor experiences, of the hidden causes that lead to the registered observations.  So, guided by evolutionarily-shaped inductive learning and information processing biases, the robots will be able to exhibit the tendency to organize the gathered experiences into event-predictive encodings.

Consequently, endowed with suitable IFOL information-processing biases, the robot's AI may develop that will be able to explain the reality it is confronted with, reason about it, and find adaptive solutions, making it Strong AI.


\bibliographystyle{IEEEbib}
\bibliography{mydb}

\end{document}